\documentclass[conference,compsoc]{IEEEtran}
\usepackage{cite} 
\usepackage{enumitem}
\ifCLASSOPTIONcompsoc
  \usepackage{cite}
\else
  \usepackage{cite}
\fi

\usepackage{graphicx}
\ifCLASSINFOpdf
\else
\fi
\begin{document}
%
\title{Artificial Intelligence-driven Intelligent Wearable Systems: \\
A full-stack Integration from Material Design to Personalized Interaction}


\author{\IEEEauthorblockN{Jingyi Zhao}
\IEEEauthorblockA{The second hospital of Jilin University \\
Northeast Asia Active Aging Laboratory\\
Jilin, China\\
Email: zhaojingyi249@163.com}

\and
\IEEEauthorblockN{Daqian Shi}
\IEEEauthorblockA{QMUL Digital Environment Research Institute \\UCL Institute Of Health Informatics, \\ Lonodon, UK \\
Email: d.shi@qmul.ac.uk}

\and
\IEEEauthorblockN{Zhengda Wang}
\IEEEauthorblockA{The second hospital of Jilin University \\
Northeast Asia Active Aging Laboratory\\
Jilin, China\\
Email: wangzd9920@mails.jlu.edu.cn}

\and
\IEEEauthorblockN{Xiongfeng Tang}
\IEEEauthorblockA{The second hospital of Jilin University \\
College of Artificial Intelligence, \\Jilin University,
Jilin, China\\
Email: tangxf921@jlu.edu.cn}

\and
\IEEEauthorblockN{Yanguo Qin}
\IEEEauthorblockA{The second hospital of Jilin University \\
Northeast Asia Active Aging Laboratory\\
Jilin, China\\
Email: qinyg@jlu.edu.cn}}


%


\maketitle

\begin{abstract}
Intelligent wearable systems are at the forefront of precision medicine and play a crucial role in enhancing human-machine interaction. Traditional devices often encounter limitations due to their dependence on empirical material design and basic signal processing techniques. To overcome these issues, we introduce the concept of Human-Symbiotic Health Intelligence (HSHI), which is a framework that integrates multi-modal sensor networks with edge–cloud collaborative computing and a hybrid approach to data and knowledge modeling. HSHI is designed to adapt dynamically to both inter-individual and intra-individual variability, transitioning health management from passive monitoring to an active collaborative evolution. The framework incorporates AI-driven optimization of materials and micro-structures, provides robust interpretation of multi-modal signals, and utilizes a dual mechanism that merges population-level insights with personalized adaptations. Moreover, the integration of closed-loop optimization through reinforcement learning and digital twins facilitates customized interventions and feedback. In general, HSHI represents a significant shift in healthcare, moving towards a model that emphasizes prevention, adaptability, and a harmonious relationship between technology and health management.
\end{abstract}


%
\IEEEpeerreviewmaketitle

\section{Introduction}
Wearable devices have emerged as a key enabler of precision medicine, ubiquitous health monitoring, and human–computer interaction. Recent advances in flexible electronics, microfluidic systems, and functional nanomaterials allow wearable sensors to continuously capture a wide spectrum of physiological and biochemical signals\cite{Gao2016, Yang2022, Manjakkal2021}. However, the design of such devices remains non-trivial: they must balance mechanical flexibility, long-term biocompatibility, and robustness under dynamic conditions, requiring iterative optimization of materials, structures, and system integration. Meanwhile, collected data exhibit distinct characteristics, high-frequency time-series, multimodal heterogeneity, and strong non-stationarity, often contaminated by noise and influenced by individual or environmental variations\cite{Sabry2022}. These properties significantly increase the complexity of downstream data processing and interpretation.

Despite progress in materials and device engineering, the conventional “synthesis–characterization–optimization” cycle for the development of wearable sensors is still labor-intensive and heavily based on trial-and-error\cite{Zhang2021}. At the data analysis level, traditional statistical tools can extract partial information, but they are insufficient for real-time, personalized, and predictive health monitoring \cite{Elfouly2025}. Artificial intelligence (AI) techniques, including deep learning, generative modeling, multi-modal learning, and edge–cloud collaborative computing, have the potential to address these limitations. AI can not only accelerate the design of sensing materials and device architectures through data-driven optimization, but also enhance the analysis of complex physiological datasets, allowing robust feature extraction, anomaly detection, and predictive feedback in real-time \cite{Ghadi2025,Ferrara2024}.

In this context, we introduce the concept of \textbf{Human-Symbiotic Health Intelligence (HSHI)}, a novel paradigm that redefines wearable systems from passive data collectors to active adaptive health agents. The core of HSHI lies in the integration of multimodal sensor networks (wearable, implantable, and contactless), edge–cloud collaborative architectures, and dual-driven models that combine population-level medical knowledge (large models) with individual-specific adaptive learning (small models). This framework enables continuous co-evolution between human health states and intelligent systems, fostering a “symbiotic” relationship rather than one-way assistance. The key contributions of this work are summarized as follows:

\begin{enumerate}[label=(\arabic*)]
    \item We propose a health intelligent agent, HSHI, based on edge-cloud collaboration and multi-modal fusion, integrating large-scale health knowledge and individual continuous data to support dynamic and personalized health modeling. 
    \item We develop an AI-driven pipeline that streamlines sensor/material design and data processing, enabling real-time prediction and closed-loop feedback.
    \item We run the proposed method on a sweat sensing, demonstrating the necessity and potential of HSHI for health monitoring in complex and dynamic environments.
\end{enumerate}


\section{The Proposed Method}
\begin{figure}
    \centering
    \includegraphics[width=1\linewidth]{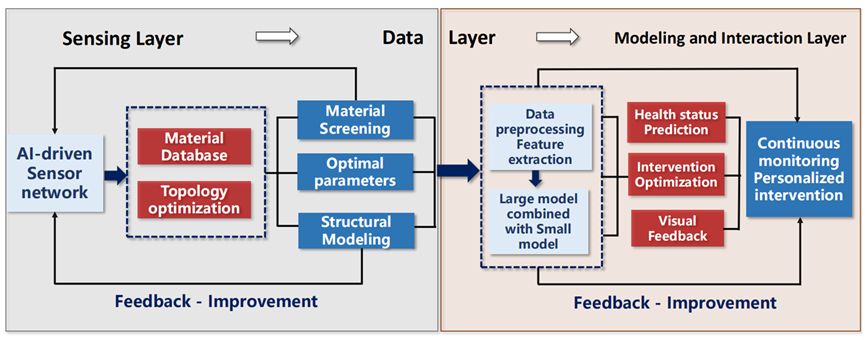}
    \caption{Framework of Human-Symbiotic Health Intelligence (HSHI)}
    \label{fig:Framework}
\end{figure}

This study proposes a novel HSHI framework, aiming to break through the current bottlenecks in material design, signal processing, and intelligent modeling of wearable health monitoring. This framework integrates the construction of sensor networks, data preprocessing, and model prediction and interaction to achieve continuous monitoring, personalized modeling, and closed-loop intervention of individual multimodal physiological signals. Figure \ref{fig:Framework} depicts the complete framework, which is organized into three tightly coupled modules: HSHI operates at three key layers:(i) Sensing Layer: HSHI utilizes material science databases and advanced topology optimization techniques to quickly screen and develop flexible sensors that cater to the needs of multi-modal monitoring, including electrical, optical, acoustic, and magnetic measurements. (ii) Data Layer: HSHI processes and extracts features from diverse types of data, including time-series, multi-modal, high-dimensional, sparse, and individual longitudinal data, creating a high-quality health database.(iii) Modeling and Interaction Layer: HSHI integrates group universal models with individualized small models to facilitate health status prediction and optimize intervention plans. This approach provides visual feedback and enhances the health management by shifting from passive monitoring to active engagement.

\subsection{Sensor Network Establishment}

In wearable health monitoring systems, selecting functional materials and optimizing micro-structures are crucial factors that influence the sensitivity, stability, and long-term reliability of sensors \cite{Park2018, Chen2023}. However, traditional research often follows an empirical iterative process of``synthesis - characterization - optimization." This approach is not only time-consuming and resource-intensive but also makes it difficult to efficiently optimize multiple goals simultaneously, such as conductivity, flexibility, and biocompatibility. This methodological challenge becomes even more pronounced in complex scenarios like sweat electrochemical sensing and optical blood oxygen monitoring.

Under the HSHI framework, researchers can input the monitoring target and constraints, and the system will conduct dual screening and optimization of materials and structures based on the material database  and combined with high-throughput modeling and generative algorithms. To be specific:

For sweat electrochemical sensors, HSHI can predict the water retention, processability, and biocompatibility of polymer and hydrogel materials based on their molecular structural characteristics. It can also identify candidate systems that exhibit both conductive stability and mechanical flexibility using graph neural networks (GNN) and generative models. Additionally, with the assistance of microfluidic simulation and topology optimization algorithms, the system can design efficient fluid channel geometries for sweat collection and transport. This improves detection sensitivity and response speed while minimizing signal attenuation caused by evaporation or diffusion \cite{Du2024}.

In the case of optical sensors, HSHI can leverage databases of photonic crystals and two-dimensional materials. By integrating deep learning and reinforcement learning methods, it can predict and optimize optical transmission characteristics across various wavelengths. This allows for the recommendation of the most suitable sensing materials for skin-tissue interface conditions. In addition, HSHI can provide designs for waveguides or electrode microstructures aimed at maximizing signal collection efficiency and improving tissue penetration depth \cite{Chen2025,Zhong2025}. Moreover, it is capable of utilizing reverse design models to automatically generate optimized optical path structures tailored to the characteristics of the target signal, which effectively reduces the costs associated with experimental trial and error.

The closed-loop mechanism of database modeling and feedback transitions the traditional reliance on experience in material and structure design to an active, data-driven, and intelligent approach. Unlike conventional iterative processes, HSHI can notably reduce the research and development cycle, minimize resource consumption, and evolve wearable sensing networks from single-point optimizations to a more collaborative global framework. This shift lays the methodological groundwork for personalized health management and enhances multimodal vital sign detection capabilities.

\subsection{Data Processing and Intelligent Modeling}

\subsubsection{Data preprocessing and Feature extraction}

In the HSHI framework, the raw data collected from the sensor network can be categorized into four types: 
(1) Time series data, including ECG, blood pressure waveforms, and respiratory signals; 
(2) Multimodal fusion data, comprising electrical, optical, and strain/mechanical signals;
(3) High-dimensional sparse data, such as sweat composition profiles and metabolomics features; and (4) Longitudinal individualized data, representing long-term continuous health trajectories. 
Each data type presents specific challenges: time series data exhibit autocorrelation and drift; multi-modal fusion data suffer from signal misalignment and cross-modal inconsistency; high-dimensional sparse data are affected by strong noise, limited sample sizes, and the curse of dimensionality; and longitudinal individualized data are characterized by substantial inter-individual variability, strong non-stationarity, and high risks of model degradation. Traditional statistical approaches, such as linear regression, multivariate analysis of variance, and principal component analysis, often fail to adequately address these nonlinear, heterogeneous, noisy, and incomplete data problems. Therefore, HSHI establishes an automated pipeline that includes data preprocessing, feature extraction, and model optimization, providing a solid foundation for subsequent health state characterization and personalized intervention.

Firstly, for high-dimensional sparse data, we use unsupervised learning algorithms to reduce dimension and compress features, through which we can map the sparse high-dimensional vectors to a low-dimensional latent space, making the subsequent modeling more robust. Secondly, for unlabeled time series and multi-modal sensing signals, we introduce the Self-Supervised Learning and Transformer encoder to generate a unified feature embedding, which retains temporal context and is compatible with cross-modal alignment. Thus, through the unsupervised and self-supervised stage, we obtain a time series consistent, modality-aligned, and denoised feature representation dataset.

Next, for the subsets with labels , we use supervised learning algorithms (Support Vector Machine, Random Forest, and XGBoost) to train classification or regression models in the embedding space to achieve discrimination or risk scoring of health status. At this point, our feature representation evolves from the unified low-dimensional space in the unlabeled stage to a labeled domain model with discriminative ability.

In scenarios with strong temporal dependencies and high demands for cross-modal fusion, HSHI employs deep learning models, such as multi-modal Transformers, temporal graph neural networks, and attention mechanism fusion modules, to integrate embedded features for cross-modal fusion modeling and individualized prediction. In addition, our goal is to incorporate a contrastive learning module into the HSHI model in the future to improve the segmentation of similar health status sample data. Following these processing steps, we will create a ``structured, low-dimensional, aligned, and applicable for prediction and intervention" health database. This database will allow us to generate a unified multi-modal feature vector for each individual at each time point. 

Using this database, we can smoothly advance towards developing universal models and personalized approaches, as well as prediction, intervention, and interaction strategies.

\subsubsection{Universal modeling and Personalized evolution}
Based on this, HSHI advances to the stage of universal modeling and personalized evolution. Using the standardized and aligned dataset obtained from the preceding preprocessing and feature extraction pipeline, the large model leverages extensive medical cohorts , multi-center databases, and cross-institutional datasets. Through knowledge graph construction and self-supervised learning, it extracts cross-modal common representations to derive population-level health patterns and predictive capabilities. Currently, the small model is deployed on edge devices and continuously adapts to individual longitudinal health data streams through incremental learning, contrastive learning, and federated transfer learning, thereby capturing individual variability and dynamic health trajectories. The large and small models engage in a mutually beneficial symbiotic relationship through a spiral iterative mechanism of “group knowledge transfer—individual feedback correction”, where the large model provides population-level patterns and strategy guidance, while the small model provides individual-specific features and boundary conditions, continuously refining the overall health model. This collaborative framework achieves hierarchical coupling from population-wide universal modeling to individualized evolution, effectively balancing model generalization with personalized adaptability.
\subsubsection{Prediction - Intervention – Interaction}
Ultimately, the objective of HSHI is to establish a closed-loop optimization framework of “prediction–intervention–interaction”, enabling proactive and individualized health management. In this framework, temporal deep learning models are used to model longitudinal health data dynamically. This module adopts a hybrid parallel structure, combining the long-range dependency modeling of TCN with the local time convolution characteristics, to achieve a balance between multi-scale feature extraction and real-time prediction performance. 
Furthermore, HSHI incorporates individualized digital twins (DTs) to construct dynamic virtual representations of individual health states, facilitating parallel simulation and risk assessment of multiple intervention plans. The results are then translated into interactive knowledge through large language models (LLMs) and explainable AI techniques, and subsequently provided to clinicians and users\cite{Shi2025a,Wu2024}. This virtual-real bidirectional mapping enhances the interpretability and trustworthiness of human–computer interaction while providing robust support for chronic disease management, precision medicine, and public health applications.
To highlight the uniqueness of the HSHI framework compared to existing research paradigms, we conducted a systematic comparative analysis (Table 1) \ref{fig:Table1}. The HSHI framework seamlessly integrates material intelligent design, edge-cloud collaborative learning, and individualized modeling mechanisms. This integration overcomes the one-dimensional optimization limitations found in traditional frameworks.
\begin{table}
    \centering
    \includegraphics[width=0.9\linewidth]{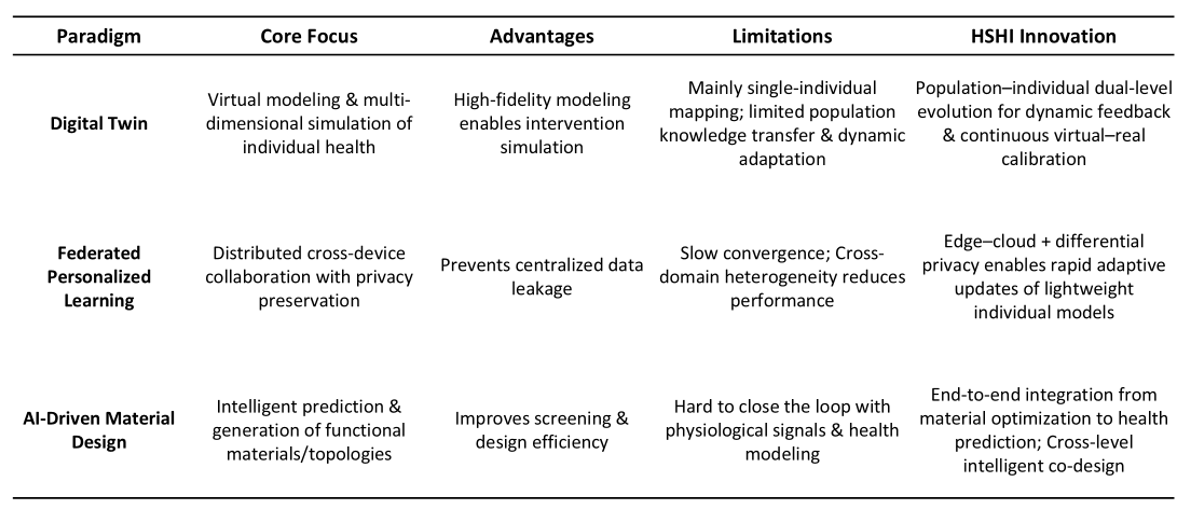}
    \caption{Comparison of Traditional Paradigms and HSHI}
    \label{fig:Table1}
\end{table}

\section{Experimental Observations}
To validate the feasibility and scientific validity of the proposed HSHI’s framework in the material, sensing, and data modeling layers, a multi-level and multimodal experimental protocol was established. The combined use of standardized characterization, physical simulation, and AI-assisted analysis was aimed at verifying the efficiency and robustness of AI-driven design in wearable health systems.

\subsection{Material Evaluation}
For the sweat sensor, we conducted a systematic characterization of candidate materials at the material level. This process included a thorough analysis from chemical structure to macroscopic performance, utilizing AI and generative algorithms to ensure that model predictions aligned with actual performance.

Chemical structure characterization was first performed to confirm molecular composition and crosslinking configurations, ensuring bio-safety and stability in physiological environments. Subsequently, micro-structural and morphological analyzes were applied to examine the capacity of the material for sweat adsorption, diffusion, and transport.

Macroscopic evaluations covered water retention and swelling behavior, mechanical and rheological properties, adhesion, and electrochemical conductivity. These assessments enabled a comprehensive evaluation of AI-based material design in achieving multi-objective optimization among flexibility, conductivity, and hydration balance.

\subsection{ Sensor Validation}
At the sensor level, a comprehensive validation scheme was implemented to assess the performance of AI-optimized architectures in electrochemical and optical sensing modules.
For the electrochemical modules, cyclic voltammetry, electrochemical impedance spectroscopy measurements were employed to analyze the coupling between electrode topology, interfacial properties, and ionic transport pathways, thus verifying the theoretical predictions on signal selectivity and response dynamics.

This layer of evaluation demonstrates that the AI-enabled multiphysics co-design paradigm can effectively guide the construction and optimization of flexible sensors with enhanced adaptability and robustness under dynamic human physiological conditions.
\subsection{Data Evaluation}
At the data-processing and modeling level, experimental assessments were conducted to verify the reliability, generalizability, and stability of the AI pipeline for health-state recognition and prediction.
Self-supervised and contrastive learning models were first evaluated to determine the temporal consistency, cross-modal compatibility, and noise robustness of learned feature representations. Standard machine-learning classifiers such as Support Vector Machine, Random Forest, and XGBoost were then trained on the extracted feature space to assess the reliability of AI-driven feature extraction and signal preprocessing.

Finally, Transformer-based temporal models and personalized adaptive learning frameworks were employed to examine whether the collaborative evolution between large-scale and lightweight models can achieve dynamic prediction and individualized adaptation. These results collectively support the validity of the HSHI data-model loop and demonstrate its capacity for long-term, personalized health inference.

\section{Conclusion and Future Work}
This paper proposes and systematically elaborates the overall framework of HSHI, aiming to promote health management from ``passive monitoring" to ``active symbiosis". HSHI integrates multi-modal sensors, edge-cloud collaborative computing, the bidirectional evolution mechanism of large and small models, and DTs-driven closed-loop intervention, achieving full-chain optimization from data collection to intelligent modeling and human-machine interaction, breaking through the limitations of existing intelligent health systems that are static, one-way, and fragmented.

In the future, HSHI is expected to develop into a long-term symbiotic partner for human life health. It can not only real-time sense the physiological state of individuals, but also gradually understand individual behavioral habits, living environment, and social psychological factors through continuous learning and iteration, thereby providing comprehensive health support. With the continuous progress of generative AI, multi-modal large models, and human-machine interaction technologies, HSHI will have the ability to adapt, self-learn, and take active interventions, gradually achieving the transition from an intelligent tool to an intelligent companion. In the long term, HSHI signifies not just a technological advancement, but also a trans-formative change in the medical paradigm transitioning from a ``disease-oriented" approach to a ``health-oriented" one, and moving from passive diagnosis and treatment to active prevention. This progression aims to foster health symbiosis at both individual and societal levels, ultimately enhancing overall well-being.





%

\end{document}